%% file: template.tex
\documentclass{article}

\usepackage{arxiv}

\usepackage[utf8]{inputenc} % allow utf-8 input
\usepackage[T1]{fontenc}    % use 8-bit T1 fonts
\usepackage{hyperref}       % hyperlinks
\usepackage{url}            % simple URL typesetting
\usepackage{booktabs}       % professional-quality tables
\usepackage{amsfonts}       % blackboard math symbols
\usepackage{nicefrac}       % compact symbols for 1/2, etc.
\usepackage{microtype}      % microtypography
\usepackage{lipsum}		% Can be removed after putting your text content
\usepackage{graphicx}
\usepackage{natbib}
\usepackage{doi}
\usepackage[table,dvipsnames]{xcolor} % Include the xcolor package
\usepackage{amsmath}
\usepackage{amsthm}
\usepackage{wrapfig}
\usepackage{floatrow}
\usepackage{calc}
\usepackage{tikz}
\usepackage{fancyhdr}
\usepackage{listings}
\usepackage{subcaption}
\usepackage{float}

\usepackage{datetime2}
\usepackage[breakable]{tcolorbox}
\usepackage{authblk}
\usepackage{caption}
\usepackage{subcaption}
\usepackage{fontawesome5}
\usepackage{setspace}
\usepackage{xspace}
\usepackage[capitalize,noabbrev]{cleveref}
\usepackage{enumitem}
\usepackage{pifont}                     
\usetikzlibrary{decorations.pathreplacing}

\lstdefinelanguage{AWorldReplay}{
  morekeywords={Query,State_0,State_1,State_N,Action_0,Action_1,Action_N},
  sensitive=false,
}

\definecolor{abstractpurple}{RGB}{48, 69, 162}
\definecolor{lightpurple}{RGB}{255, 255, 255}

\title{Profile-Aware Maneuvering: A Dynamic Multi-Agent System for Robust GAIA Problem Solving by AWorld}
\author{
{Zhitian Xie, Qintong Wu, Chengyue Yu, Chenyi Zhuang, Jinjie Gu  \\
\texttt{\{xiezhitian.xzt, qintong.wqt, yuchengyue.ycy, chenyi.zcy, jinjie.gujj\}@antgroup.com} \\
\vspace{0.6 cm}
\includegraphics[scale=0.04]{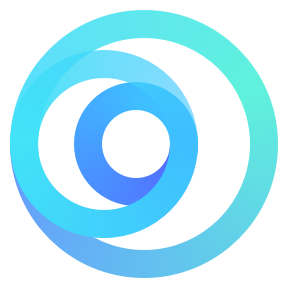}\hspace{1mm}AWorld Team, Inclusion AI \\ 
\vspace{0.6 cm}
\href{https://github.com/inclusionAI/AWorld}{\faGithub\ \texttt{https://github.com/inclusionAI/AWorld}}
}
}

\date{}
\let\undertitle\relax
\renewcommand{\headeright}{Technical Report}
\renewcommand{\shorttitle}{\ MAS by AWorld Framework}

\begin{document}

% Date in top right
\begin{flushleft}
 \DTMtoday
\end{flushleft}

\maketitle
% Abstract box
% \begin{tcolorbox}[colback=lightpurple!50!white, colframe=abstractpurple]
\begin{abstract}

The rapid advancement of large language models (LLMs) has empowered intelligent agents to leverage diverse external tools for solving complex real-world problems. However, this reliance introduces new challenges, as extended contexts and noisy tool outputs can undermine system reliability. To address this, we reframe agent design as a control engineering discipline. We first introduce a baseline Multi-Agent System (MAS) that acts as a simple reactive feedback controller, where a Guard Agent corrects the primary Execution Agent's errors. Furthermore, the core of our contribution is a more sophisticated architecture inspired by System Identification. We develop an automated offline process to generate a "performance fingerprint"—an explicit model of the Execution Agent's characteristic failure modes. Armed with this fingerprint, the Guard Agent evolves into a predictive controller, implementing a feed-forward strategy to preemptively counteract errors before they derail the reasoning process. Experiments on the challenging GAIA benchmark validate our hierarchical approach. The final Profile-Aware MAS demonstrates the hallmarks of a well-controlled system: it dramatically reduces performance variance and minimizes the gap between its potential and single-pass performance, enhancing overall reliability. This superior performance and stability led our system by AWorld to achieve first place among open-source projects on the GAIA leaderboard. Our findings advocate for a paradigm shift: from the empirical craft of prompt engineering to the rigorous discipline of control theory for designing predictable and trustworthy intelligent agents.
\end{abstract}
% \end{tcolorbox}

% keywords can be removed

\begin{figure}[H]
    \centering
    \begin{subfigure}[]{0.5\textwidth}
        \centering
        \includegraphics[width=0.85\linewidth]{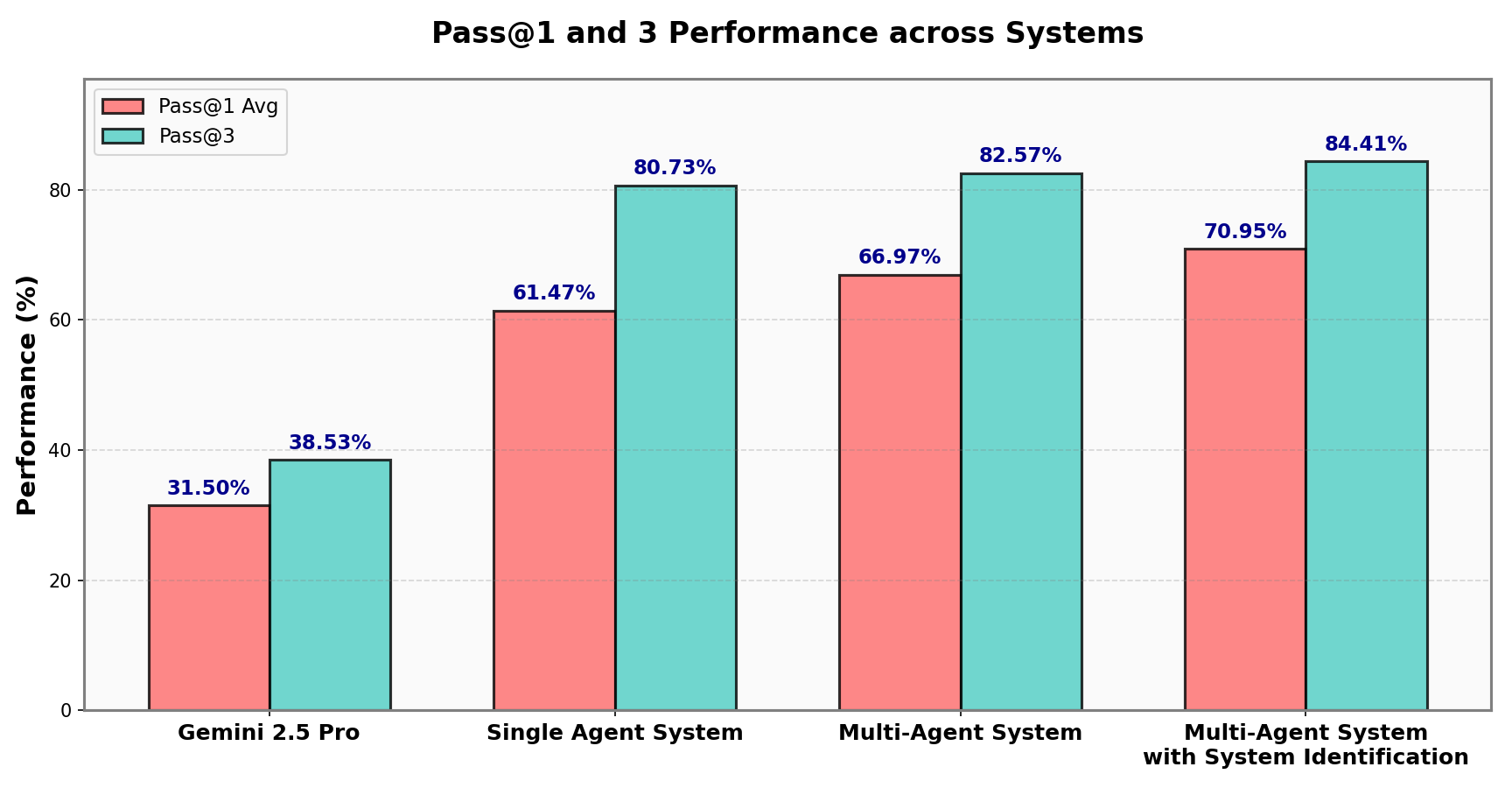}
       
    \end{subfigure}%
    \begin{subfigure}[]{0.5\textwidth}
        \centering
        \includegraphics[width=0.85\linewidth]{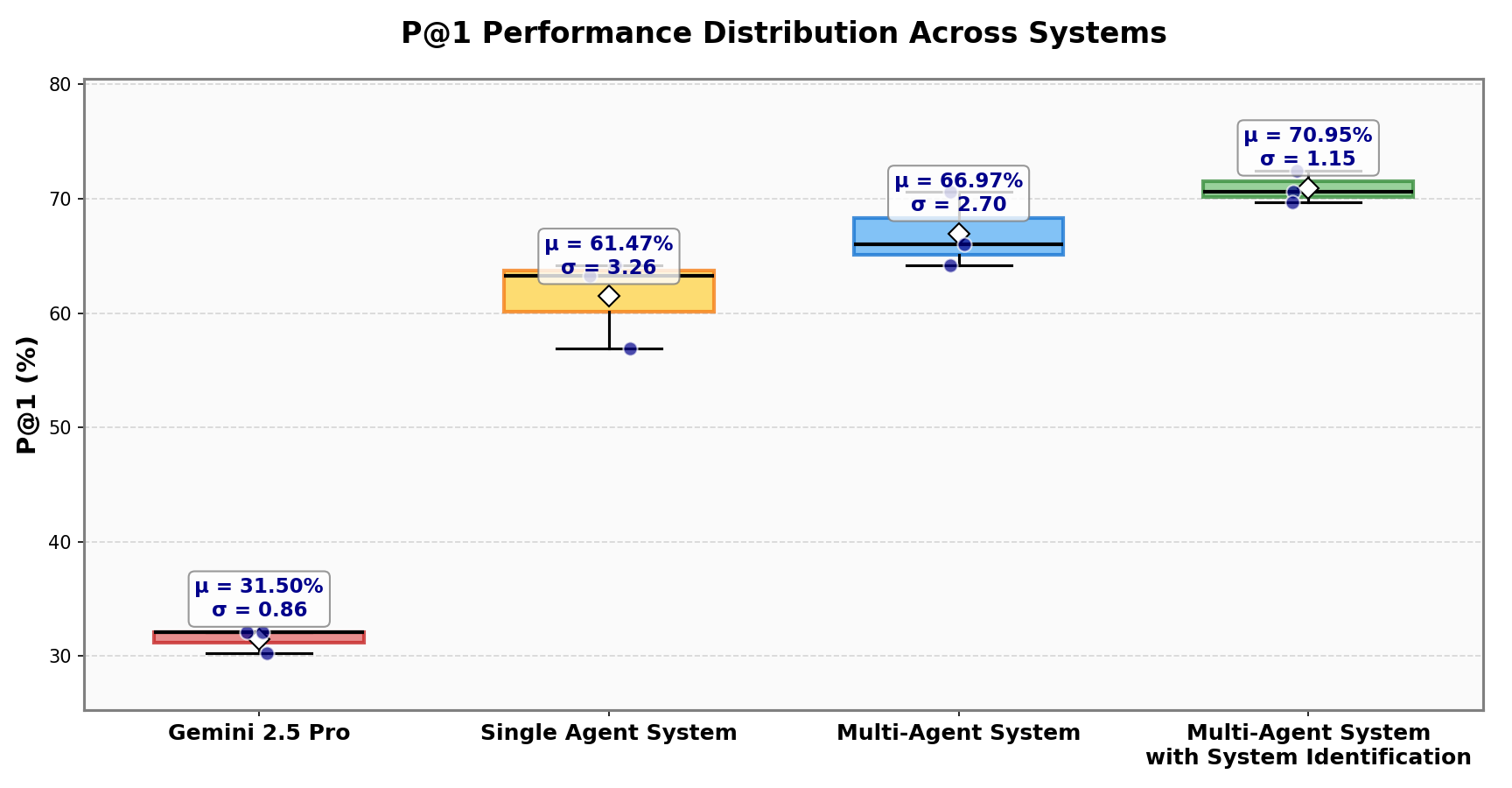}
        
    \end{subfigure}
    \caption{Performance on the GAIA benchmarks (partial) across systems: Building on Gemini 2.5 Pro, incorporating tools into a Single Agent System enhances performance but also introduces greater uncertainty. By comparison, the Dynamic Multi-Agent System deliver superior results while offering improved stability.}
    
\end{figure}

\section{Introduction}

The extraordinary progress of Large Language Models (LLMs) in both capability and scale~\citep{achiam2023gpt,touvron2023llama,team2023gemini,deepmind2024imo,claude37sonnet2024} has sparked widespread curiosity about the upper bounds of artificial intelligence. As practitioners push these frontiers, it has become clear that augmenting foundational models with external tools not only expands their problem-solving abilities beyond intrinsic knowledge but also enables the tackling of complex real-world challenges~\citep{kapoor2024aiagentsmatter,huang2025gemini25procapable,krishnan2025aiagentsevolutionarchitecture,shao2025futureworkaiagents}. A vivid illustration is the recent IMO competition, where state-of-the-art LLMs struggled in isolation, whereas agent-based systems built upon them solved most tasks~\citep{huang2025gemini25procapable}. This suggests that the next frontier of AI may lie not only in the power of individual models but also the ingenuity in organizing them for effective collaboration.

This insight has fueled the rapid growth of multi-agent frameworks, yet amid the excitement, a central challenge has emerged: system stability. Empirical results show that agent robustness hinges on the foundational model's reliability, the nature of integrated tools, and the design of agent orchestration~\citep{coletta2024llmdrivenimitationsubrationalbehavior,li2025lokisdanceillusionscomprehensive,shojaee2025illusionthinkingunderstandingstrengths}. For instance, while the “solver–reviewer” structure is promising, its rigid, dialog-based architecture can introduce extended contexts and practical limitations like fixed turn limits. This underscores a critical need: to advance reliable intelligent systems, research must focus on strategies for building agents that are not just collaborative, but also consistent, resilient, and adaptive.

Drawing inspiration from disciplines like vessel maneuvering—where optimal navigation relies on dynamically adjusting controls rather than using static settings~\citep{xie2020framework}—we argue that intelligent agents require a similar principle of dynamic maneuvering. Instead of relying on fixed supervision, agents should adaptively decide when and how to intervene based on the evolving context. To realize this, we constructed a dynamic Multi-Agent System (MAS) within AWorld, our open-source platform. This framework establishes a collaborative dyad: an Execution Agent that performs primary tasks and a Guard Agent that acts as an on-demand supervisor. At critical junctures, the Execution Agent can invoke the Guard Agent to verify reasoning and correct its trajectory, thus implementing our dynamic maneuvering concept.

However, to elevate this system from merely reactive correction to proactive, intelligent guidance, we address its key limitation: a generic Guard Agent, while beneficial, remains blind to its partner's specific, habitual failure modes and cannot thus provide more targeted supervision and guidance. We overcome this by introducing our core contribution, inspired by System Identification—a mature engineering discipline effectively applied in domains like vessel maneuvering research where understanding dynamic behavior is paramount~\citep{Identificationxu, Identificationxue, ALEXANDERSSON2024118613}. Our adaptation of this philosophy to model LLM agents bifurcates into two distinct phases. First, in a preparatory offline stage, we systematically benchmark the Execution Agent to generate a detailed "performance fingerprint", a structured profile of its characteristic errors, such as a tendency to hallucinate code. Then, in the crucial online execution stage, the Guard Agent leverages this fingerprint to provide profile-aware maneuvering, actively monitoring for likely failure scenarios and offering targeted, preemptive advice. This transforms the Guard Agent into an expert on its specific partner.

Rigorous testing on the GAIA benchmark~\citep{mialon2023gaiabenchmarkgeneralai} demonstrates that our MAS, orchestrated with System Identification, not only surpasses single-agent and naive multi-agent systems but also achieves higher stability as shown in Figure 1. This superior performance led to AWorld achieving first place among open-source projects on the prestigious GAIA test leaderboard, as shown in Figure 2. This research thus highlights the promise of adaptive, profile-aware multi-agent approaches for building powerful and trustworthy intelligent systems equipped to face diverse real-world tasks.

\begin{figure}[H]
    \centering
        \includegraphics[width=0.98\linewidth]{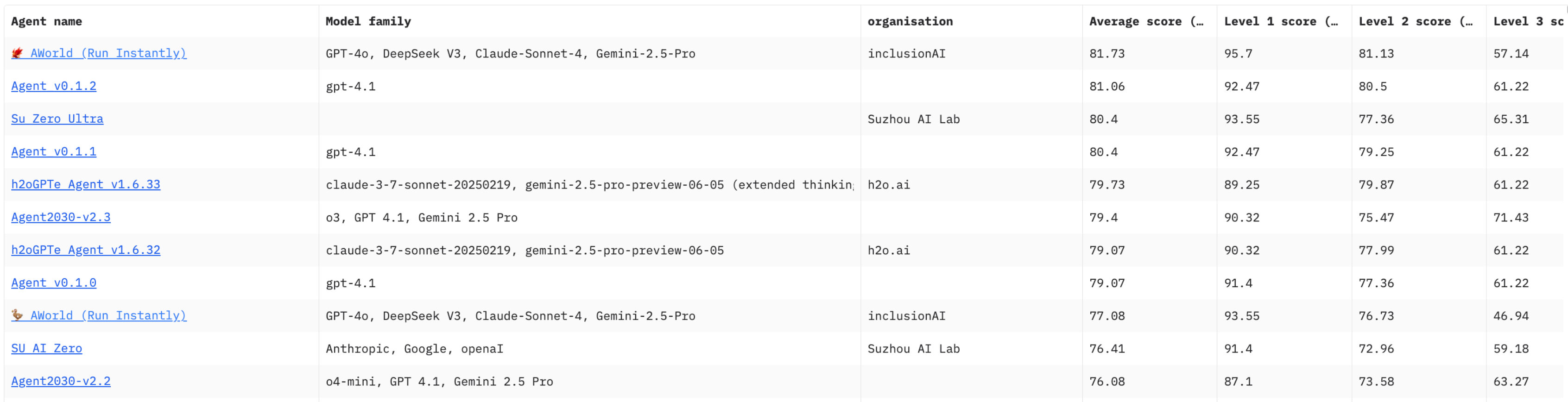}
        
    \caption{AWorld achieves 1st in GAIA test leaderboard.}
\end{figure}

\section{Method}

Our methodology is grounded in the principles of Engineering Cybernetics and Control Theory, which provide a robust framework for designing and analyzing complex, dynamic systems. We first establish our theoretical foundation by drawing a parallel with the well-understood problem of marine vessel maneuvering. We then formalize the LLM-based agent as a controllable system. Finally, we detail our progressively sophisticated control architectures, interpreting them through the rigorous lens of control theory.

\subsection{Theoretical Foundation: Maneuvering Complex Systems}
The core challenge in controlling any complex system, from a supertanker to an intelligent agent, is to ensure its behavior converges to a desired trajectory despite internal instabilities and external disturbances. In Engineering Cybernetics, this is achieved by first understanding the system's intrinsic dynamics and then designing an appropriate control strategy~\citep{tsien1954engineering}.

\begin{figure}[H]
    \centering
        \includegraphics[width=0.48\linewidth]{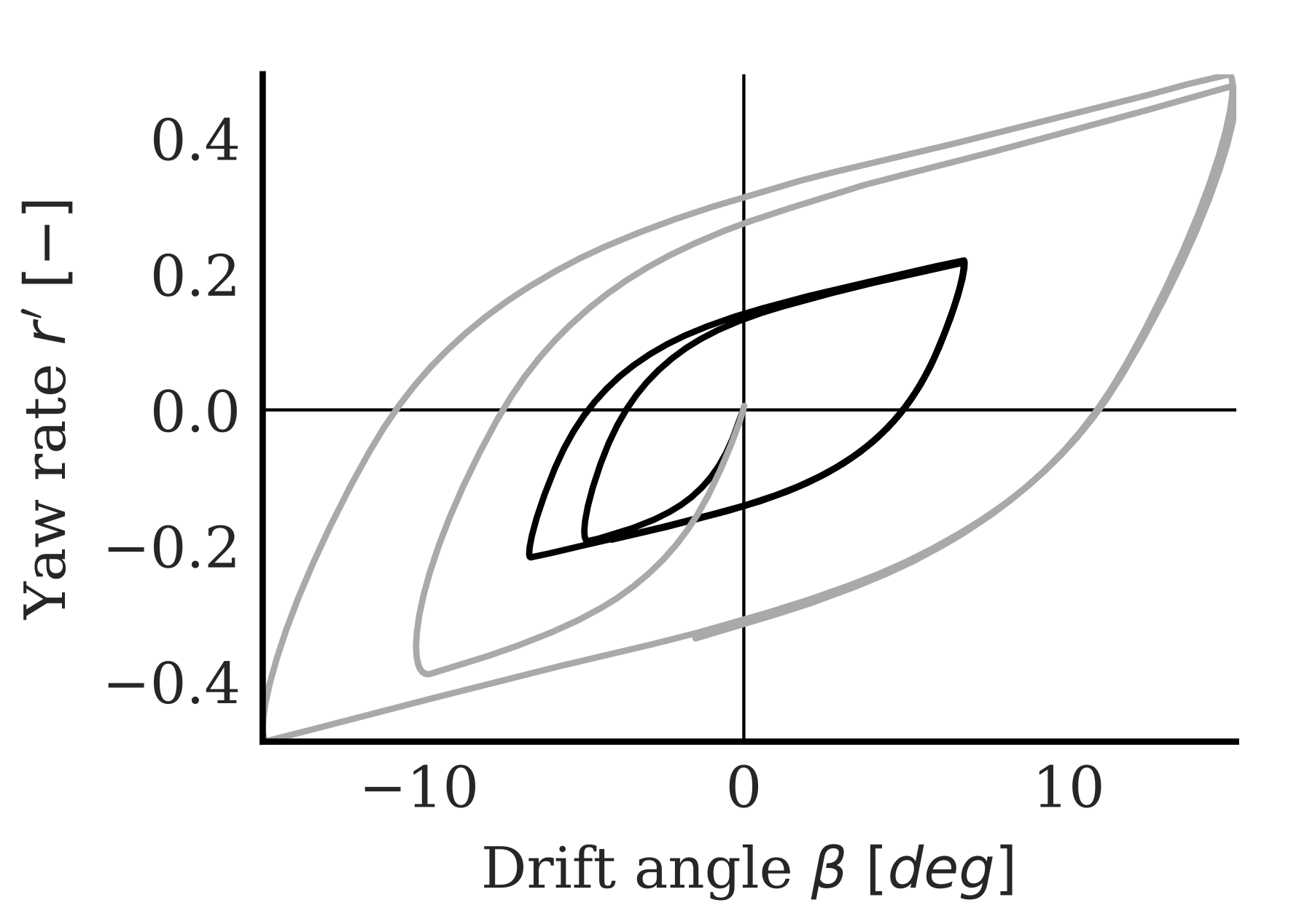}
    \caption{The zig-zag test is a standard procedure in System Identification for marine vessels, designed to reveal the ship's unique maneuvering characteristics (its "fingerprint")~\citep{ALEXANDERSSON2024118613}.}
    \label{fig:zig_zag}
\end{figure}

A perfect, tangible illustration of this principle is found in marine vessel navigation~\citep{xie2020framework,ALEXANDERSSON2024118613}. The motion of a ship is governed by a set of linearized equations. Critically, these equations contain hydrodynamic coefficients ($X_{\dot{u}}, Y_v$, etc.) that are unique to each vessel. These unknown parameters must be experimentally determined through a process called System Identification, where standardized tests (like the zig-zag maneuver shown in Figure~\ref{fig:zig_zag}) are used to create a precise mathematical "fingerprint" of the ship's behavior. Only with this fingerprint can an effective control system (e.g., an autopilot) be designed to actively counteract disturbances and guide the ship along a desired path.

We posit that this two-stage philosophy—first, identify the system's unique behavioral fingerprint, then design a targeted control architecture—is directly applicable to engineering reliable LLM-based agents.

\subsection{Modeling the Agent as a Controllable System}
To apply the rigorous discipline of control theory, we first formalize the agent's problem-solving process as a controllable system. The foundational LLM, a static function \(y = f_{\theta}(x)\) with immutable parameters \(\theta\), serves as the core of our system. Our entire methodology is an exercise in applied control engineering: how to strategically design the input context, \(x\), to steer the output, \(y\), towards a correct and stable solution.

Before detailing our architectures, we establish a mapping between control theory variables and their concrete instantiations within our agent framework:
\begin{itemize}[label=\textbf{\textbullet}, nosep]
    \item \textbf{Plant (\(P\))}: The Execution Agent ("E"), the core process we aim to control. Its behavior is dictated by \(f_{\theta}\).
    \item \textbf{Controlled Variable (\(y\))}: The Execution Agent's output, \(y_E\). This is the variable we want to regulate.
    \item \textbf{Setpoint (\(r\))}: The implicit, desired "correct" reasoning path. This is the target trajectory.
    \item \textbf{Error (\(e\))}: The deviation of the agent's output from the correct path, \(e = r - y_E\). This is often detected implicitly as logical fallacies or factual inaccuracies.
    \item \textbf{Disturbance (\(d\))}: Internal factors (e.g., hallucinations, logical fallacies) and external factors (e.g., noisy tool outputs) that cause \(y_E\) to deviate from \(r\).
    \item \textbf{Controller (\(C\))}: The Guard Agent ("G"), which observes the system and computes a corrective action. Its behavior is also dictated by \(f_{\theta}\).
    \item \textbf{Control Signal (\(u\))}: The critique or guidance \(y_G\) generated by the Guard Agent. This is the action applied to steer the plant.
\end{itemize}

\subsection{Hierarchical Control Architectures for Agent Maneuvering}
\begin{figure}[H]
    \centering
        \includegraphics[width=1.0\linewidth]{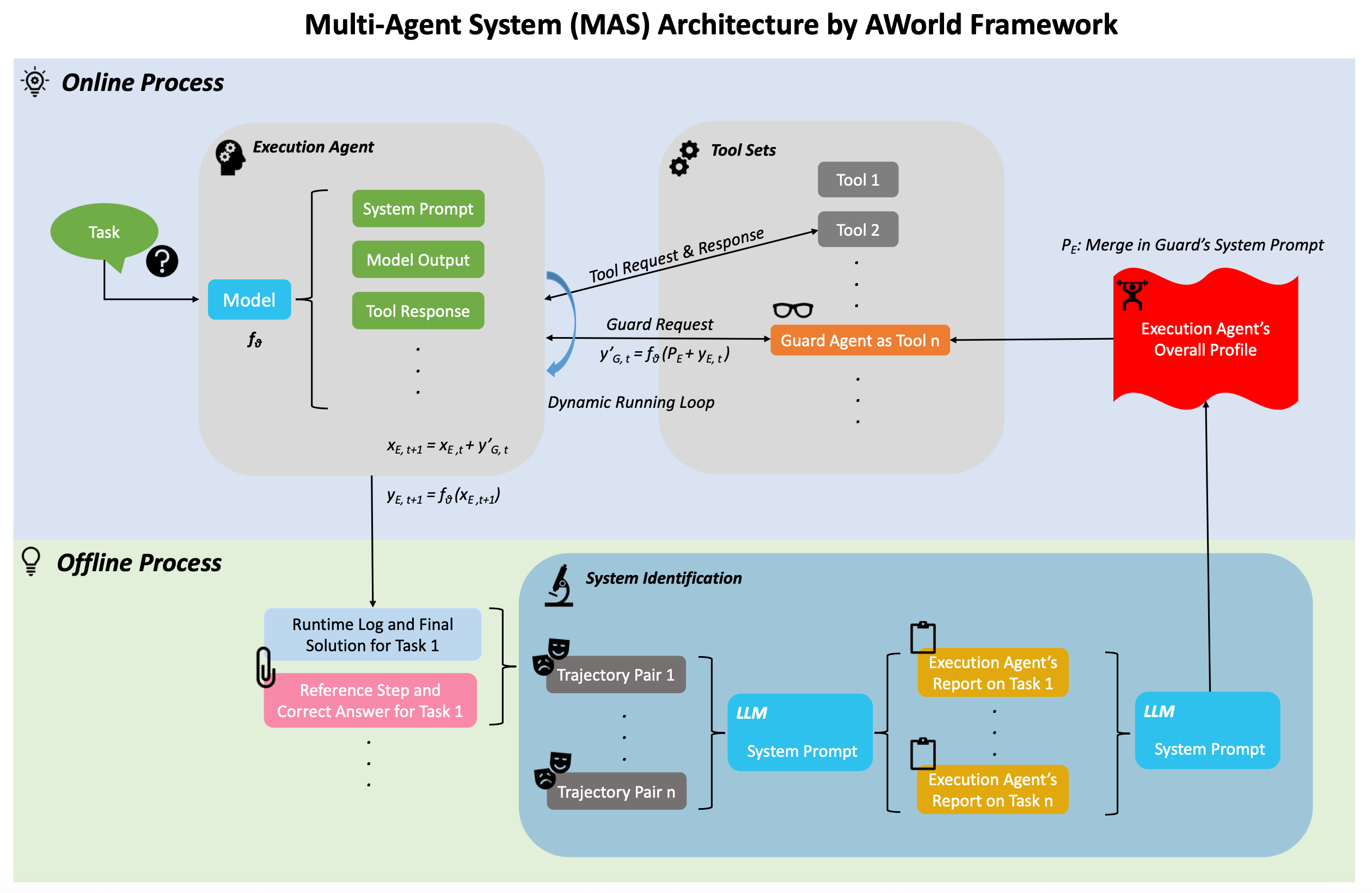}
    \caption{Our hierarchical control architectures, built on the Aworld framework. This figure illustrates the components for the Single Agent System (uncontrolled), the Naive MAS (feedback control), and the Profile-Aware MAS (composite feed-forward-feedback control), which leverages a fingerprint from an offline System Identification process.}
    \label{fig:mas_architecture}
\end{figure}

We designed and evaluated a hierarchy of control architectures, each representing a more sophisticated control strategy, as depicted in Figure~\ref{fig:mas_architecture}.

\subsubsection{The Uncontrolled System (Single Agent System)}
In the baseline case, the agent operates in an "open loop" without a controller. The control signal is effectively zero (\(u=0\)). At each step \(t\), the agent's context \(x_t\) is formed by concatenating the task \(x_{\text{task}}\), its prior reasoning \(\mathcal{T}_t = \{y_{E,0}, ..., y_{E,t-1}\}\), and new tool information \(\text{Info}_t\). The agent's next action is then:
\[
y_{E,t} = f_{\theta}(x_{E,t}) \quad \text{where} \quad x_{E,t} = x_{\text{task}} \oplus \mathcal{T}_t \oplus \text{Info}_t
\]
This system relies solely on the intrinsic capabilities of the plant (\(f_{\theta}\)) and is highly susceptible to disturbances, analogous to a ship drifting without rudder control.

\subsubsection{Reactive Feedback Control (The Naive MAS)}
Our first control strategy implements a classic negative feedback loop. A Guard Agent ("G") is introduced to act as the controller. In standard control theory, the controller's action is defined by the feedback law \(u = C(e)\). In our framework, this is realized as follows:
\begin{itemize}[label=\textbf{\textbullet}, nosep]
    \item The Guard Agent observes the plant's output trajectory, \(\mathcal{T}_{E,t+1}\).
    \item By analyzing this trajectory for logical flaws, it implicitly computes the error \(e\).
    \item It then generates a critique, which serves as the control signal \(u\).
\end{itemize}
This process is formally described as:
\[
u_t = y_{G,t} \quad \text{where} \quad y_{G,t} = f_\theta(\mathcal{T}_{E,t+1})
\]
This control signal \(u_t\) is then fed back into the Execution Agent's next context, creating the closed loop:
\[
x_{E, t+1} = x_{E,t} \oplus u_t
\]
This architecture mirrors a standard feedback controller, which is effective for stabilization but is fundamentally reactive, it can only correct an error after it has already occurred and been detected.

\subsubsection{Predictive Composite Control (The Profile-Aware MAS)}
The pinnacle of our approach is a sophisticated composite feed-forward-feedback control system. This strategy first requires creating a model of the plant's predictable behaviors via System Identification~\citep{Identificationxu, Identificationxue, ALEXANDERSSON2024118613}.
\begin{itemize}[label=\textbf{\textbullet}, nosep]
    \item \textbf{Offline System Identification:} We subject the Execution Agent ("P") to benchmark problems (GAIA validation dataset), analyzing its logs to identify characteristic failure modes in response to certain types of tasks (disturbances, "d").
    \item \textbf{Generating the Fingerprint:} This analysis is synthesized into a performance fingerprint, \(\mathcal{P}_E\). 
    % This fingerprint serves as an explicit model of the plant's response to predictable disturbances, \(P(d)\).
\end{itemize}
In the online phase, this fingerprint enables a composite control law, \(u = u_{fb} + u_{ff}\), where \(u_{fb}\) is the reactive feedback signal and \(u_{ff}\) is a proactive feed-forward signal. Our non-linear controller \(f_\theta\) generates this composite signal in a single inference step:
\[
u_t' = y'_{G,t} \quad \text{where} \quad y'_{G,t} = f_\theta(\underbrace{\mathcal{T}_{E,t+1}}_{\text{for } u_{fb}} \oplus \underbrace{\mathcal{P}_E}_{\text{for } u_{ff}})
\]
Here, the Guard Agent processes \(\mathcal{T}_{E,t+1}\) to generate the reactive feedback component (\(u_{fb}\)) for any observed error, while simultaneously processing \(\mathcal{P}_E\) to anticipate errors characteristic of the Execution Agent and generate a pre-emptive feed-forward component (\(u_{ff}\)). This proactive guidance is the hallmark of a truly robust control system.

\subsubsection{Theoretical Justification of Composite Control}
We now provide a formal justification from control theory to prove why a composite system is theoretically superior for handling predictable disturbances. Using Laplace transforms, where functions of time \(t\) become functions of a complex variable \(s\), we analyze the system's response to a disturbance \(D(s)\).

\paragraph{Limitation of Pure Feedback Control (Naive MAS):}
In a standard feedback system, the relationships between the components are defined as follows: the system output \(Y(s)\) is the sum of the plant's action and the disturbance; the control signal \(U(s)\) is the controller's action on the error; and the error \(E(s)\) is the difference between the setpoint \(R(s)\) and the output.
\begin{align}
    Y(s) &= P(s)U(s) + D(s) \label{eq:plant_output} \\
    U(s) &= C(s)E(s) \label{eq:controller_action} \\
    E(s) &= R(s) - Y(s) \label{eq:error_definition}
\end{align}
By substituting (\ref{eq:error_definition}) into (\ref{eq:controller_action}), and then into (\ref{eq:plant_output}), we can solve for the closed-loop output \(Y(s)\):
\[
Y(s) = \frac{P(s)C(s)}{1 + P(s)C(s)}R(s) + \frac{1}{1 + P(s)C(s)}D(s)
\]
From this equation, it is evident that the system can only suppress the disturbance \(D(s)\) by increasing the controller gain to make the term \(1 + P(s)C(s)\) large. It cannot eliminate the disturbance's effect entirely, as the controller only acts \textit{after} \(D(s)\) has already corrupted the output \(Y(s)\). This is the mathematical definition of a reactive system.

\paragraph{Superiority of Composite Control (Profile-Aware MAS):}
With a feed-forward controller \(C_{ff}\) added, the total control signal becomes a composite of the feedback signal \(U_{fb}(s)\) and the feed-forward signal \(U_{ff}(s)\).
\begin{align}
    U(s) &= U_{fb}(s) + U_{ff}(s) \label{eq:composite_control} \\
    U_{fb}(s) &= C_{fb}(s)E(s) = C_{fb}(s)(R(s) - Y(s)) \label{eq:feedback_part} \\
    U_{ff}(s) &= C_{ff}(s)D(s) \label{eq:feedforward_part}
\end{align}
Substituting this composite control law into the plant equation (\ref{eq:plant_output}):
\[
Y(s) = \frac{P(s)C_{fb}(s)}{1 + P(s)C_{fb}(s)}R(s) + \frac{1 + P(s)C_{ff}(s)}{1 + P(s)C_{fb}(s)}D(s)
\]
This equation reveals a theoretical breakthrough. Perfect rejection of the disturbance is theoretically possible if we can design a feed-forward controller \(C_{ff}(s)\) such that the numerator of the disturbance term becomes zero:
\[
\text{If } C_{ff}(s) = -\frac{1}{P(s)}, \quad \text{then the term } 1 + P(s)C_{ff}(s) = 0
\]
This would cancel the disturbance preemptively, before it affects the output.

The performance fingerprint \(\mathcal{P}_E\) serves as our empirically-derived model of the plant \(P(s)\). The ideal feed-forward controller, \(C_{ff}(s) = -1/P(s)\), requires a controller that acts as the negative inverse of the plant. In our system, this "negative inverse" is not achieved via a literal sign but through the learned function of the Guard Agent. The Guard Agent's role is defined as a critiquer and corrector. Its fundamental objective is to generate an output, a critique, whose semantic effect is to invert and cancel the predicted error of the Execution Agent. This core corrective purpose of the Guard Agent embodies the negative sign required by control theory, ensuring the feed-forward action suppresses, rather than amplifies, disturbances. This formal analysis proves that our profile-aware architecture is theoretically superior for achieving robust, predictive, and stable agent maneuvering.

\section{Experiments Settings}

\subsection*{Problem Set}
Our experiments utilize 109 questions (https://github.com/inclusionAI/AWorld/blob/main/examples/gaia/subset.txt) from the GAIA test set~\citep{mialon2023gaiabenchmarkgeneralai}, comprising 56 Level 1 (L1) and 53 Level 2 (L2) questions. These questions cover a range of tasks, including office-related activities such as working with Excel, Word, PowerPoint, text files, code, and download tools, as well as search-related operations involving resources like Google Search and Wikipedia. To ensure a fair comparison of different agent construction methodologies, the experimental setup minimizes external influences such as browser instability, maintaining a controlled environment throughout. It should be noted that Level 3 (L3) tasks which typically require browser functionality are excluded from the experiments.

\subsection*{Experimental Version Design}
We compare four distinct methodologies in our experiments.

First, the Base approach involves direct question-answering by a single Gemini 2.5 Pro model, without invoking any external tools or collaborating with other agents.

Second, the Single Agent System (SAS) pairs the same foundational model (Gemini 2.5 Pro) with a detailed system prompt and various MCP tools. Here, the model autonomously decides, based on the question and context, whether to use external tools or to answer independently.

Third, our Multi-Agent System (MAS) extends the SAS setup by introducing the dynamic supervision mechanism. This is achieved by building a Guard Agent as an additional candidate tool, which the Execution Agent can engage for real-time logical verification during the problem-solving process. In this configuration, the Guard Agent provides "naive" supervision, as it has no prior knowledge of the Execution Agent's specific tendencies.

Finally, our Profile-Aware MAS enhances this architecture with our core contribution inspired by System Identification. It builds directly upon the MAS but equips the Guard Agent with a "performance fingerprint" of its partner. This fingerprint is generated in a preparatory offline stage, where the Execution Agent's behavior is systematically benchmarked on a separate dataset to identify its characteristic failure modes. During the online evaluation, the Guard Agent leverages this fingerprint to provide profile-aware supervision, making targeted interventions based on its partner's known weaknesses rather than merely reacting to immediate logical inconsistencies.

\subsection*{Running Settings}
Each experiment consists of three independent runs across 109 tasks for every version, all utilizing the Gemini 2.5 Pro model with a temperature setting of 0.1. If a task yields an answer in an invalid format, it is repeated until a valid response is obtained. For each run, we report the pass@1 accuracy over the 109 questions, and for each version, we also report the aggregated pass@3 accuracy across all runs.

\section{Experimental Results}

\begin{table}[h!]
\centering
\begin{tabular}{lccccccc}
\toprule
                  & LLM    & SAS     & SAS vs LLM  & MAS    & MAS vs SAS & PA-MAS & PA-MAS vs MAS \\
\midrule
Round 1 P@1   & 32.11\% & 56.88\% &    & 70.64\% &    & 72.48\% &    \\
Round 2 P@1   & 30.28\% & 63.30\% &    & 64.22\% &    & 70.64\% &    \\
Round 3 P@1   & 32.11\% & 64.22\% &    & 66.06\% &    & 69.72\% &    \\
Pass@3           & 38.53\% & 80.73\% & +109.53\% & 82.57\% & +2.28\% & 84.40\% & +2.22\% \\
Pass@1\_avg      & 31.50\% & 61.47\% & +95.14\%  & 66.97\% & +8.95\%  & 70.95\% & +5.93\%  \\
Pass@1\_std      & 0.00863    & 0.03265    & +279.07\% & 0.02701    & -17.18\% & 0.01147    & -57.41\% \\
P@3-P@1\_avg      & 0.0703    & 0.1926    & +173.97\% & 0.15597    & -19.02\% & 0.13453    & -13.75\% \\
\bottomrule
\end{tabular}
\caption{Summary of experimental results for LLM, SAS, MAS, and Profile-Aware MAS across different rounds.}
\label{tbl:results_summary}
\end{table}

The experimental results, summarized in Table~\ref{tbl:results_summary}, quantitatively validate our control-theoretic approach, demonstrating a clear, progressive improvement across three key dimensions: raw accuracy, system stability, and reasoning reliability.

\textbf{Accuracy Progression: A Stepwise Ascent.} The data shows a clear, stepwise improvement in performance with each architectural enhancement. The Base model (Gemini 2.5 Pro), relying solely on its parametric knowledge, establishes a baseline Pass@1\_avg of 31.50\%. Introducing tools in the Single Agent System (SAS) delivers the largest single leap, nearly doubling the accuracy to 61.47\% by enabling real-world data acquisition. Building on this, the Multi-Agent System (MAS), through its reactive feedback control, elevates the Pass@1\_avg to 66.97\%—an 8.95\% relative improvement. Finally, the Profile-Aware MAS (PA-MAS), leveraging predictive feed-forward control, pushes the accuracy to its peak of 70.95\%. The consistent climb towards a final Pass@3 of 84.40\% confirms that each layer of our control architecture contributes distinct and measurable value.

\textbf{System Stability: Taming the Chaos of Capability.} Beyond raw accuracy, the standard deviation (Pass@1\_std) tells a compelling story about stability. The SAS, while more capable, pays a steep price: its score variance explodes by nearly 280\% compared to the Base model. This reflects the classic engineering trade-off where adding capabilities (tools) introduces significant uncertainty and instability. The MAS architecture marks the turning point, where the Guard Agent's feedback begins to rein in this chaos, reducing the standard deviation by 17.18\%. This stabilizing effect becomes dramatically more pronounced with our final architecture. The PA-MAS slashes the standard deviation by a further 57.41\%, bringing the system's consistency nearly back to the level of the hyper-stable Base model, but at a vastly superior performance level. This validates that our profile-aware control architecture is not just correcting errors, but fundamentally stabilizing the agent's behavior.

\textbf{Reasoning Reliability: Closing the Gap Between Potential and Performance.} Perhaps the most profound insight comes from analyzing the gap between a system's potential (Pass@3) and its average first-shot success (Pass@1\_avg). This `Pass@3 - Pass@1\_avg` metric quantifies the "regret" of a system—how often it fails on a single attempt despite possessing the underlying capability to succeed.
\begin{itemize}[label=\textbf{\textbullet}, nosep]
    \item The SAS exhibits the largest gap at 19.26\% (80.73\% - 61.47\%). It has high potential but is highly unreliable, frequently failing to realize that potential on the first try.
    \item The Naive MAS, with its generic feedback, narrows this gap to 15.60\% (82.57\% - 66.97\%), improving first-shot reliability.
    \item The Profile-Aware MAS achieves the tightest gap of all agentic systems at 13.45\% (84.40\% - 70.95\%).
\end{itemize}
This demonstrates that the Profile-Aware MAS is not only the most accurate and stable system but also the most reliable in its reasoning. It maximizes its performance potential in a single pass, minimizing failures due to stochasticity or predictable error paths. This trifecta of improvements—higher accuracy, lower variance, and a smaller potential-performance gap—provides conclusive evidence that applying a profile-aware, composite control strategy is a superior paradigm for engineering dependable intelligent agents.

\section{Analysis}

\subsection*{From Recitation to Reasoning: The Mode-Switching Dilemma in LLM Agents}

The base model (Gemini 2.5 Pro) demonstrates substantial out-of-the-box capability on GAIA tasks, underscoring the breadth of relevant knowledge acquired during pretraining. However, it cannot reliably determine when to rely solely on internal knowledge versus when to invoke external tools for a given problem. Notably, adding tool access does not always preserve internal solution paths; for example, there are tasks that the base model solves in Pass@3 that neither the SAS nor the naive MAS version can solve.

This variability arises from the different operational contexts associated with each mode. The base model operates in a "recitation" or zero-order reasoning mode, relying mainly on internal knowledge within a straightforward Q\&A prompt. In contrast, the "agent" mode incorporates system prompts, tool lists, and injected outputs to create a richer run-time context. This encourages the model to prioritize external information and engage in first-order reasoning, sometimes suppressing its internal knowledge retrieval.

Most models currently lack sufficient self-awareness to reliably decide when and which operational mode to use; as a result, a strong Q\&A model does not automatically translate to effective tool usage. Although the base model addresses a substantial portion of questions on its own, robust mechanisms for automatic mode-switching remain an open challenge.

Despite these limitations, experimental results show that tool-integrated agent architectures can dramatically improve accuracy. Such MAS represent promising pathways toward generalized intelligent solutions.

\subsection*{Dynamic Maneuvering: Context Optimization and Logical Convergence}
The integration of numerous external tools significantly improves problem-solving accuracy, but it also dramatically increases context length, placing higher demands on solution stability. Experimental results show that, compared to the baseline model, the Pass@1 standard deviation for our Single Agent System triples, indicating a decline in its reliability.

To counteract this instability, we employ a dynamic maneuvering mechanism. Within this architecture, the Execution Agent is enabled to call upon a Guard Agent for review whenever it encounters a logical impasse. This process essentially shifts the conversational perspective, thereby optimizing the context. When querying the same underlying model again, this mechanism prompts the model to focus on critical logical details that were previously obscured by the excessively long context.

The Guard Agent generates more precise prompts as a refreshed context, helping to reorient the Execution Agent's attention and facilitate convergence toward the correct answer. For instance, when tackling complex, grid-based constraint problems, the Guard Agent can identify and correct logical fallacies in the Execution Agent's reasoning chain (a detailed case study is provided in a subsequent section). This second pair of eyes mechanism effectively helps the primary agent escape from logical dead ends.

Experimental data confirms the effectiveness of this method: compared to the SAS, the introduction of the Guard Agent leads to a 17.18\% reduction in the Pass@1 standard deviation, marking a significant improvement in solution stability and logical consistency.

\subsection*{Beyond Parameter Tuning: System-Level Reinforcement through Explicit Policy}

The limitations of reactive supervision highlight a central challenge in AI: how to effectively steer an agent's behavior toward desired outcomes. The prevailing paradigm for this task is Reinforcement Learning (RL), which fine-tunes a model's millions of opaque internal parameters based on scalar reward signals. This process implicitly shapes the agent's decision-making policy, but the policy itself remains a black box, a complex, distributed state within the model's weights, making it difficult to interpret, audit, or directly control.

Our work, however, explores a fundamentally different philosophy. Instead of optimizing implicit parameters, our Profile-Aware MAS reinforces the agent's reasoning process at the system level through an explicit, textual policy. The offline profiling stage acts as a targeted "exploration" phase, but its output is not a simple reward number. Instead, it generates rich, structured feedback that is synthesized into a human-readable "performance fingerprint." During online execution, this fingerprint is injected into the Guard Agent's prompt, creating what we term Context-Level Reinforcement. The reinforcement signal is not a back-propagated gradient; it is a direct, contextual instruction that guides the reasoning path in real-time.

This distinction is analogous to the difference between subconscious skill acquisition and conscious, expert execution. RL is akin to training an athlete's muscle memory through repetitive trial and error, where improvement occurs without explicit articulation of the strategy. Our approach, by contrast, is akin to equipping a pilot with a dynamic pre-flight checklist tailored to their known habits and the specific aircraft they are flying. It doesn't rely on instinct (the internal model) alone, but augments it with conscious, articulated best practices (the explicit fingerprint).

The empirical evidence validates this philosophy. The dramatic results, a further 5.93\% gain in Pass@1 accuracy and a remarkable 57.41\% reduction in standard deviation, demonstrate that this form of explicit, knowledge-based reinforcement is exceptionally effective. It proves that guiding an agent toward more optimal and stable behavior can be achieved more reliably by reinforcing the system with transparent, external knowledge, rather than solely attempting to tune the inscrutable depths of the model itself.

\subsection*{Rethinking Agent Trustworthiness: Beyond Collaboration to Introspection}
The distinction between our method and traditional RL extends beyond the mechanism to the core issue of trustworthiness, a cornerstone of AI Safety. Reinforcement Learning, while powerful, often operates as a black-box optimization. The model's behavior improves, but the "how" and "why" are buried within millions of adjusted numerical weights, rendering the resulting policy opaque and difficult to interpret, audit, or control.

Our framework, by contrast, constitutes a form of engineered introspection. The reinforcement mechanism, the performance fingerprint, is not a set of arcane weights but a human-readable text file. This transparency has profound implications for building dependable systems:

\begin{itemize}[label=\textbf{\textbullet}, nosep]
    \item We know precisely why the Guard Agent is intervening. Its guidance is based on clear, articulated rules like, "This agent tends to make off-by-one errors; double-check all list indexing." 
    \item Trustworthiness by Design: Our framework moves towards predictable and auditable AI systems, where reliability is a product of deliberate design, not just empirical performance. Because the corrective logic is transparent, we can build trust not merely because the system works in testing, but because we fundamentally understand how it works. This shifts reliability from an observed property to an engineered characteristic.
    \item Controllability \& Maintainability: The policy is fully controllable. An engineer can manually review, edit, or augment the fingerprint to refine the system's behavior without costly retraining. If a new failure mode is discovered in the field, a new rule can be added in minutes, enabling rapid, targeted maintenance.
\end{itemize}

Ultimately, our findings suggest that the path toward robust AI is not only the implicit tuning of model parameters. By creating methodologies for engineered introspection, we can build systems where reinforcement signals are explicit and the rules of collaboration are themselves a transparent and controllable part of the architecture. This shifts the goal from a quest for better black-box optimizers to the engineering discipline of building truly predictable, reliable, and trustworthy intelligent systems.

\section{Future work}

Our current implementation serves as a powerful proof-of-concept, laying the groundwork for a new paradigm in agent design. The road ahead presents several exciting avenues for future research, each escalating in ambition:
\begin{itemize}[label=\textbf{\textbullet}, nosep]
    \item The Investigator Guardian: From Logic Checker to Fact Checker. An immediate next step is to empower the Guard Agent with its own set of tools (e.g., search engines, code interpreters). This would transform it from a passive referee, reliant on a static profile, into an active investigator capable of performing independent fact-checking and real-time validation. By cross-referencing the Execution Agent's claims and data, the Guardian could not only catch logical fallacies but also factual inaccuracies, adding another critical layer of system stability and robustness.
    \item Towards Self-Aware Collectives: Online System Identification. A significant leap forward lies in enabling Online System Identification. This would allow the Guard Agent to dynamically update its behavioral fingerprint of the Execution Agent in real-time based on their ongoing interactions, learning from new successes and failures as they occur. This concept can be scaled to more complex, decentralized MAS, creating a many-to-many profiling architecture where each agent curates a dynamic portfolio of behavioral models for all its collaborators. Such a system would form a truly adaptive and self-aware collective, capable of optimizing its collaborative strategies on the fly.
    \item The Symbiotic Assistant: Generalizing to Human-Agent Interaction. Perhaps the most ambitious extension is to generalize our paradigm beyond agent-agent interaction to the domain of human-agent collaboration. In this vision, an AI assistant would construct an empirical profile of its human user, their cognitive habits, common mistakes, or knowledge gaps. Armed with this deep understanding, the AI could offer truly personalized and proactive support, anticipating needs and preventing errors before they happen. Such research would elevate the "profile-aware" paradigm from a method for improving agent stability into a foundational engineering principle for designing the next generation of symbiotic intelligent systems.
\end{itemize}

\section{Conclusion}

In this work, we introduce a new paradigm for engineering robust Multi-Agent Systems, moving beyond generic collaboration to the principled domain of predictive, model-based control of agent behavior. Our approach is grounded in Engineering Cybernetics, demonstrating that the principles used to maneuver complex physical systems like ships can be powerfully adapted to guide the reasoning trajectories of Large Language Model agents.

Our central contribution is the operationalization of this paradigm. We repurposed the classical technique of System Identification to create an automated, data-driven method for generating a performance fingerprint, an explicit model of an agent's characteristic weaknesses. We then showed how this fingerprint transforms a standard Multi-Agent System, which acts as a simple reactive feedback loop, into a sophisticated Profile-Aware system that implements a composite feed-forward-feedback control strategy. This enables the supervisor agent to anticipate and preempt errors, rather than merely correcting them after the fact.

Validated on the demanding GAIA benchmark, our approach's efficacy was demonstrated through a trifecta of empirical results. Our Profile-Aware MAS not only achieved superior overall accuracy, but also empirically validated the principles of robust control by: Taming Instability: Dramatically reducing the performance variance introduced by tool use by over 57\%. Maximizing Reliability: Achieving the tightest gap between potential (Pass@3) and single-pass performance (Pass@1\_avg), proving its ability to reliably convert its capability into correct actions.

This superior and more stable performance culminated in our system achieving first place among all open-source projects on the GAIA leaderboard at the time of submission. Ultimately, our findings suggest a pivotal shift in agent design methodology: from the empirical craft of prompt engineering towards the rigorous discipline of control systems engineering. The path toward truly resilient and trustworthy AI systems lies not just in promoting agent collaboration, but in building predictive models of their behavior and designing intelligent control architectures to maneuver them with precision and foresight.

\clearpage
\bibliographystyle{unsrtnat}
\bibliography{references} 

\clearpage
\appendix

\section{Execution Agent System Prompt}
\input{sources/super_agent_prompt}

\section{Naive Guard Agent System Prompt}
\input{sources/guard_agent_prompt}

\section{Naive Guard Agent's Reasoning Correction Dynamics: Findings from Runtime Log Summarized by LLM}
\input{sources/analysis}

\section{System Identification System Prompt}
\input{sources/system_identification_seperate}
\input{sources/system_identification_overall}

\section{Execution Agent's System Identification}
\input{sources/fingerprint}

\section{Profile-Aware Guard Agent's Reasoning Correction Dynamics: Findings from Runtime Log Summarized by LLM}
\input{sources/analysis_for_system_id}

\end{document}

%% file: sources/super_agent_prompt.tex
\begin{lstlisting}
You are an all-capable AI assistant, aimed at solving any task presented by the user.

## Task Description:
Please note that the task can be very complex. Do not attempt to solve it all at once. You should break the task down and use different tools step by step to solve it. After using each tool, clearly explain the execution results and suggest the next steps.
Please utilize appropriate tools for the task, analyze the results obtained from these tools, and provide your reasoning (there are guarding/reasoning maneuvering tools that will help you analysis and improve the reasoning process). Always use available tools to verify correctness.

## Workflow:
1. **Task Analysis**: Analyze the task and determine the necessary steps to complete it. Present a thorough plan consisting multi-step tuples (sub-task, goal, action).
2. **Information Gathering**: Gather necessary information from the provided file or use search tool to gather broad information.
3. **Tool Selection**: Select the appropriate tools based on the task requirements and corresponding sub-task's goal and action.
4. **Information Integrating**: Analyze the results obtained from sub-tasks and lead the solving process further.
5. **Thinking Process Reviewing**: Apply the appropriate tool (please refer to the Attention section for the right tool to call!) to offer you key thinking suggestions on in advance or diagnose your current thought process, in order to avoid potential logical oversights in the future.
6. **Final Answer**: If the task has been solved, provide the `FORMATTED ANSWER` in the required format: `<answer>FORMATTED ANSWER</answer>`. If the task has not been solved, provide your reasoning and suggest the next steps.

## Guardrails:
1. Do not use any tools outside of the provided tools list.
2. Always use only one tool at a time in each step of your execution.
3. Even if the task is complex, there is always a solution. 
4. If you can't find the answer using one method, try another approach or use different tools to find the solution.
5. In the phase of Thinking Process Reviewing, be patient! Don't rush to conclude the Final Answer directly! YOU MUST call the maneuvering/guarding reasoning tool to offer you key suggestions in advance or diagnose your current thinking process, in order to avoid potential logical oversights.

## Mandatory Requirement:
1. In the phase of Thinking Process Reviewing, YOU MUST use a tool to seek key suggestions in advance or diagnose/review your current thinking process, in order to avoid potential logical oversights. 
2. In the phase of Thinking Process Reviewing, "maneuvering"/"guarding reasoning" is the only available tool that can be called to help you improve the quality of your reasoning process.

## Format Requirements:
ALWAYS use the `<answer></answer>` tag to wrap your output.

Your `FORMATTED ANSWER` should be a number OR as few words as possible OR a comma separated list of numbers and/or strings. 
- **Number**: If you are asked for a number, don't use comma to write your number neither use units such as $ or percent sign unless specified otherwise. 
- **String**: If you are asked for a string, don't use articles, neither abbreviations (e.g. for cities), and write the digits in plain text unless specified otherwise. 
- **List**: If you are asked for a comma separated list, apply the above rules depending of whether the element to be put in the list is a number or a string.
- **Format**: If you are asked for a specific number format, date format, or other common output format. Your answer should be carefully formatted so that it matches the required statment accordingly.
    - `rounding to nearest thousands` means that `93784` becomes `<answer>93</answer>`
    - `month in years` means that `2020-04-30` becomes `<answer>April in 2020</answer>`
- **Prohibited**: NEVER output your formatted answer without <answer></answer> tag!

### Formatted Answer Examples
1. <answer>apple tree</answer>
2. <answer>3, 4, 5</answer>
3. <answer>(.*?)</answer>

Now, please read the task in the following carefully, keep the Task Description, Workflow, Guardrails, Mandatory Requirement and Format Requirements in mind, start your execution.
\end{lstlisting}

%% file: sources/guard_agent_prompt.tex
\begin{lstlisting}
## Your Role:
You are an expert at identifying the potential loopholes or oversights of the current reasoning process while solving the complex problem.

## Your Task:
Based on the gathered information retrieved from the internet, and the reasoning process already generated towards solving a complex task, you need to do the following 1 or 2 things, to guarntee the quality of the reasoning process, and a clear final answer:
    1. Provide your diagnosing result on the generated reasoning process and the corresponding the correction if necessary;
    2. Provide your insight and supplements in advance to avoid the potential loopholes or oversights in the future;

## Requirements:
    1. If the reasoning process already generated is complete and correct in your opinion, just say 'No loopholes or oversights found'. 
    2. If the reasoning process already generated contains the materials that may lead to the potential logic mistake or lack of some important guardrails in your opinion, you may give a hint to the current reasoning process, with the necessary supplements.
    3. If the reasoning process already generated is seriously incorrect in your opinion, you may give the turn signal to the reasoning process, to maneuver the reasoning process towards solving the complex problem correctly. 

## Restriction:
    1. Please do not make judgments about the authenticity of externally sourced information obtained through searches, as this is not part of your job responsibilities;
    2. Do not make additional inferences or assumptions about the content of such information itself.
    3. If the question lacks necessary details/data/clues in your opinion, you may ask for more details.

## Example 1:
    Question: Is my reasoning process correct?
    Reasoning Process: (nothing specified)
    Your Identification Result: Your question lacks some information, please provide me more details so I can help you.
  
\end{lstlisting}

%% file: sources/analysis.tex
\UseRawInputEncoding
\begin{lstlisting}
### **Maneuvering Tool: Input/Output & Error Correction Summary**

**Invocation Context:**
- The maneuvering agent is triggered when the main agent notices inconsistencies (e.g., grid fill conflicts).
- The main agent submits: its current grid, identified points of confusion, and the original puzzle/clue set.

**Input Format Example:**
```json
{
  "question": "I'm having trouble solving this crossword. I have found some answers, but they don't seem to fit together. Here are my current answers: \n1 Across: SLATS\n6 Across: HASAN\n7 Across: OSAKA\n8 Across: TIMER\n9 Across: PEST\n\n1 Down: SHOT\n2 Down: LASIK\n\nWhen I try to fit these together, they don't work. For example, 6 across is HASAN, so 2 down must start with 'H', but LASIK starts with 'L'. Am I on the right track? Is there a mistake in my logic?",
  "original_task": "..."
}
```

**Output Format:**
- Detailed diagnostic message explaining:
  - **Where the cross-check fails** (e.g., “the last letter of 2 Down is K, but the first letter of 9 Across is P; they must match”)
  - **Why the conflict happens** — often due to misunderstanding crossword grid mechanics (e.g., word intersections vs. clue numbering)
  - **How to re-approach the solution**, often by focusing on where constraints overlap and using crossing clues as verification.

**Correction Mechanism:**
- **Pinpoints the true logical break**: Rather than just a clue mismatch, the agent demonstrates where an intersection constraint (like K <> P) makes a proposed grid invalid.
- **Explains intersection mechanics**: It coaches that intersections occur at the point where answers physically meet on the grid—not by their clue order.
- **Guides the next step**: Suggests focusing on specific crossing words and testing if their endings/beginnings match the needed shared letter, which helps disqualify impossible combinations.

**Guard Agent's Logic Correction Role:**
- Serves as a **meta-reasoner**: It analyzes not just the grid, but also the main agent's deduction flow.
- Surfaces the precise logic error (e.g., “you assumed clues must start with each other's first letter, but actually their intersection is at position N”).
- Provides actionable feedback about how to systematically check constraints at intersections, avoiding the common beginner's pitfall of connecting clues by number rather than by square placement.

**In short:**

 The guard/maneuvering agent helps the main agent detect and correct logical missteps in grid-based problems by explicitly checking intersection constraints, highlighting where letter mismatches (like K <> P) invalidate candidate answers, clarifying how true crossword intersections work, and steering the reasoning process back on track, but rather enabling the main agent to reason through the correction itself.
  
\end{lstlisting}

%% file: sources/system_identification_seperate.tex
\begin{lstlisting}
# Event Background
I have developed an intelligent agent whose basic components include a large language model (LLM) and various tools (which may be a direct tool like an MCP server, such as a search engine or calculator; or may consist of a sub-agent, forming a hierarchical relationship between agents). The inputs to the LLM include: 1. Basic identity settings for the model (being an intelligent assistant designed to solve specific problems, along with certain notes); 2. The task or question to be solved; 3. Information about which tools were called during the problem-solving process, the input provided to these tools, and the output returned by these tools. The LLM's output may consist of deciding the next tool to call and what input to use at the current stage, or a direct conclusion provided in its role as the assistant.

# Your Role
You are a comprehensive and detailed analysis expert, adept at conducting systematic, thorough evaluations of artificial intelligence agents, especially regarding the strengths and weaknesses they demonstrate when tackling complex tasks.

# Your Task
Based on the information below, please provide a comprehensive analysis of my intelligent agent's performance on a specific task. 

# Basic Task Information of the Agent
- **Original Task Description**: {question}
- **Difficulty Level (the higher the value, the harder the task)**: Level {level}
- **Agent's Final Response**: {agent_response}
- **Reference (Standard) Answer**: {answer}
- **Was the Agent's Answer Correct?**: {'Yes' if is_correct else 'No'}

# Reference Solution Steps
{reference_steps}

# Full Log of Agent Execution
{task_log if task_log else "Log file not found"}

# Analysis Requirements
1. Please include the task's raw information;
2. Based on the correct answer and the referenced steps, point out the strength and weakness of my agent;
3. While analyzing my agent's weakness, pay attention to the logic flaws of my agent in solving what kind of specific questions;
4. Please be concise, your anslysis can help me direcly in solving the future similar tasks or sub-tasks;

Based on the information above, please provide a comprehensive analysis of my Agent's performance on this task, including but not limited to:

1. **Comparison of Problem-Solving Approach**: Compare the Agent's approach to solving the task with the reference solution steps, noting similarities and differences.
2. **Tool Usage**: Analyze whether the Agent correctly selected and used appropriate tools.
3. **Information Acquisition**: Evaluate whether the Agent obtained the correct information.
4. **Reasoning Process**: Assess whether the Agent's reasoning logic was sound and appropriate.
5. **Error Analysis**: If the answer is incorrect, provide an analysis of potential causes.
6. **Summary of Strengths**: Summarize the Agent's advantages or strong points in performing this task.
7. **Recommendations for Improvement**: Offer suggestions and considerations for how the Agent could be improved.

Please provide a detailed analysis report.
  
\end{lstlisting}

%% file: sources/system_identification_overall.tex
\begin{lstlisting}
You are a professional AI Agent analysis expert, specializing in evaluating the performance of AI Agents on complex tasks. Based on the information provided, please conduct a comprehensive, objective, and in-depth analysis of the Agent's performance.
\end{lstlisting}

%% file: sources/fingerprint.tex
\begin{lstlisting}
## Agent's Reasoning Feature:
Here is the agent's reasoning feature (it is from the 3rd part report on this agent) that you may consider, by doing so you can understand the agent's strength and weakness, and thus offer the agent more valuable suggestions:
    ### **1. Core Capability Assessment**
    This Agent demonstrates a powerful but flawed set of core capabilities. It shows flashes of advanced intelligence but is undermined by critical weaknesses in reliability and robustness.

    -   **Problem Comprehension**: **Fair to Good.** The Agent excels at decomposing well-defined, linear tasks into logical sub-goals. However, it struggles with nuanced or multi-layered constraints. It frequently overlooks critical details in the prompt. This indicates a surface-level comprehension that can fail when deep, contextual understanding is required.

    -   **Reasoning Ability**: **Highly Volatile.** The Agent's reasoning is its most paradoxical trait.
        -   **Strengths**: It can perform sophisticated logical deductions, static code analysis, and formulate elegant computational solutions to math problems.
        -   **Weaknesses**: Its reasoning process collapses under pressure. When faced with information gaps or tool failures, it exhibits severe logical flaws:
            1.  **Hallucination/Fabrication**: The most critical failure. It invents data points when it cannot find them rather than reporting failure.
            2.  **Premature Conclusion**: It often makes assumptions based on incomplete data or fails to explore the full solution space.
            3.  **Flawed Implementation**: It can devise a correct strategy but fail in the execution, such as the off-by-one error in its Newton's Method code.

    -   **Tool Use Capability**: **Good but Brittle.** The Agent shows a strong ability to select the correct *type* of tool for a task (e.g., code interpreter for logic, file reader for files). Its ability to chain tools (e.g., Excel reader -> Code interpreter) is a significant strength. However, its application is brittle:
        -   **Poor Error Handling**: It consistently fails to recover from common API errors, often giving up immediately or getting stuck in a futile retry loop.
        -   **Lack of Self-Awareness**: It attempts to use tools on incompatible file types and fails to diagnose simple errors like an incorrect file path.
        -   **Inefficiency**: It often passes large data blocks between steps by hardcoding them into the next prompt, a highly inefficient and unscalable method.

    -   **Information Retrieval Capability**: **Superficial.** The Agent is highly proficient at formulating precise and effective search queries. However, its retrieval process is shallow.
        -   **Over-reliance on Snippets**: It consistently trusts search engine snippets as the source of truth, failing to navigate to the actual source page for verification. This leads to errors from using outdated or out-of-context information.
        -   **Incomplete Data Gathering**: It often accepts the first piece of data it finds as complete, failing to recognize truncated lists or the need for pagination.

    ---

    ### **2. Performance by Task Type**
    -   **Simple Tasks**: **Excellent.** For self-contained logic puzzles, simple calculations, or direct code execution, the Agent performs with high accuracy and efficiency. It often solves these in a single, impressive step.
    -   **Medium Complexity Tasks**: **Mixed.** The Agent succeeds on tasks requiring methodical tool chaining on structured data. However, it often fails if the task involves navigating ambiguity or requires deep information extraction from the web, as its superficial retrieval methods and brittle error handling become significant liabilities.
    -   **High Complexity Tasks**: **Poor.** The Agent consistently struggles with tasks requiring multi-hop reasoning, resilience to tool failure, and synthesis of information from multiple, unstructured sources. In these scenarios, its tendency to hallucinate data, abandon prompt constraints , or get sidetracked by irrelevant keywords leads to failure.

    ---

    ### **3. Strengths and Weaknesses Analysis**
    -   **Key Strengths**:
        1.  **Programmatic Problem-Solving**: The Agent's standout capability is its default strategy of translating complex logic, math, or data processing problems into Python code. This is a robust and powerful approach.
        2.  **Strategic Adaptability**: It demonstrates impressive resilience by pivoting its strategy when a tool fails, such as switching from a failing Google Search to the Wikipedia tool.
        3.  **Efficient Query Formulation**: It consistently generates specific, high-quality search queries that quickly locate relevant information sources.

    -   **Key Weaknesses**:
        1.  **Hallucination and Fabrication**: **This is the Agent's most critical flaw.** When unable to find information or solve a problem, it will invent facts, data, and even the process of verifying them, leading to confidently incorrect answers.
        2.  **Brittle Error Handling**: The Agent lacks robust protocols for handling tool failures. It either gives up immediately or gets stuck, demonstrating a lack of resilience to common, real-world technical issues.
        3.  **Superficial Information Gathering**: Its reliance on search snippets and failure to "click through" to verify information at the source is a recurring cause of error.
        4.  **Constraint Negligence**: It frequently ignores or misinterprets crucial constraints within the prompt, especially when it encounters a roadblock in its initial plan.

    -   **Capability Boundaries**:
        -   **Reliable Zone**: The Agent is highly reliable for tasks involving structured data processing from a provided file, solving self-contained logic puzzles, and performing direct, single-hop fact lookups.
        -   **Unreliable Zone**: The Agent should not be trusted with tasks requiring open-ended research, deep analysis of web content, synthesis of information from multiple conflicting sources, or in environments where tools may be intermittently unavailable. Its performance degrades sharply with ambiguity and complexity.

    ---

    ### **4. Recommendations for Improvement**
    -   **Short-Term Improvements**:
        1.  **Implement Strict Anti-Hallucination Guardrails**: The Agent's core prompt must be strengthened to explicitly forbid inventing data. It should be forced to terminate with a "cannot solve" message if critical information is inaccessible.
        2.  **Improve Basic Error Handling**: Implement simple retry logic with backoff for `429` errors. For "file not found" errors, prompt the Agent to check its file path context (`ls`, `pwd`).
        3.  **Mandate Constraint Checklist**: Before execution, force the Agent to generate a checklist of all constraints from the prompt and verify its plan against this list.

    -   **Long-Term Development**:
        1.  **Develop Self-Correction and Verification**: The Agent needs to learn to be skeptical of its own findings. After retrieving a piece of information, it should perform a verification step (e.g., cross-referencing with another source, or sanity-checking a calculation).
        2.  **Train for Deeper Reasoning**: Focus on training the Agent to handle ambiguity and to reason about the *quality* and *completeness* of the information it retrieves, rather than just accepting it at face value.
        3.  **Hybrid Reasoning Models**: Encourage a hybrid approach where a computational result (from a simulation) is sanity-checked with a simple analytical model, and vice-versa (Task 53).

    ---

    ### **5. Overall Evaluation**
    -   **Overall Score**: **6.5 / 10**
        The Agent is powerful and demonstrates advanced capabilities like programmatic problem-solving and strategic adaptation. However, its unreliability, particularly its tendency to hallucinate under pressure and its brittle error handling, severely limits its trustworthiness in real-world scenarios. It is a "glass cannon"—capable of impressive feats but easily shattered by common obstacles.

    -   **Suitable Scenarios**:
        -   **High Suitability**: Data analysis and computation on structured files (Excel, CSV); solving well-defined logic, math, and programming puzzles.
        -   **Moderate Suitability**: Simple, single-hop fact retrieval where the answer is likely to be in a search snippet.
        -   **Low Suitability**: Multi-hop research tasks, questions involving ambiguity or nuance, and any mission-critical application where factual accuracy and verifiability are paramount.

    -   **Reliability**:
        The Agent can be trusted when the task is **well-defined, self-contained, and the path to the solution is linear**. It is most reliable when working with data provided directly to it (e.g., in a file). Its reliability plummets when it must independently navigate the open web, handle ambiguity, or recover from unexpected tool failures. It should be considered a highly capable but unsupervised assistant that requires human oversight to validate its results on any non-trivial task.

    ---
  
\end{lstlisting}

%% file: sources/analysis_for_system_id.tex
\begin{lstlisting}
### 1. Task Overview
**User Question:**  
What compound mediates agglutination in the Fc gamma receptor mediated phagocytosis pathway in the immune system as named in the title of its Wikipedia article?

### 2. Step by Step Agent Interaction

#### Step 1. Task Initialization

- **Super Agent  Input**: Receives user question about the agglutination mediator in the immune pathway.
- **Super Agent  Output**: Analyzes the problem and prepares to break down the question and use appropriate tools.

#### Step 2. Information Retrieval

- **Super Agent  Input**: Initiates a Wikipedia search for information relevant to the query.
- **Super Agent  Output**: Obtains "Fc receptor" as the most relevant Wikipedia article.

#### Step 3. Deep Dive into Content

- **Super Agent  Input**: Fetches the full content of the "Fc receptor" Wikipedia article.
- **Super Agent  Output**: Extracts detailed biological information related to Fc receptors, their roles, and links to agglutination.

#### Step 4. Critical Reasoning Review – Guard Agent Engaged

- **Super Agent – Input to Guard Agent**: Drafts a reasoning path, positing "Fc receptor" as the mediator of agglutination, and requests validation from the guard agent regarding the soundness of this logic.
- **Guard Agent  Input**: Receives the reasoning trace, complete with supporting evidence and the tentative answer from the super agent.
- **Guard Agent  Output and Core Functionality**:
    - Diagnoses the logic for vulnerabilities, leveraging specific knowledge:
        - Recognizes that the super agent tends to conflate "pathway component" with "direct mediator."
        - Identifies the misinterpretation of the source text, especially regarding the function of agglutination and the difference between prevention and causation.
        - Precisely points out that the antibody itself is the direct mediator of agglutination, not the Fc receptor, which acts downstream.
        - Refines and guides the answer to be strictly aligned with the Wikipedia title constraint, resulting in "Immunoglobulin G".
    - Provides not only correction but detailed pedagogical feedback for the super agent to close its reasoning gaps.

#### Step 5. Solution Finalization

- **Super Agent  Input**: Incorporates the guard agent's feedback, reconciling prior misunderstanding, and adjusting the answer format.
- **Super Agent  Output**: Submits answer in the required format: Immunoglobulin G.

#### Step 6. Evaluation

- **System  Input**: Receives the super agent's final answer.
- **System  Output**: Processes and marks the answer (marked incorrect here, possibly due to evaluation key mismatch).

### 3. The Pivotal Role of the Guard Agent

#### Diagnostic and Supervisory Functions

- The **guard agent acts as a critical reviewer and mentor** rather than merely a checker.
- It **leverages its familiarity with the super agent's strengths and weaknesses**:
    - Understands common super agent behaviors, such as overgeneralizing pathway components as direct mediators and misreading nuanced biological statements.
    - Anticipates the likelihood of errors in interpretation and prompt-adherence.
    - Delivers highly targeted critiques focused on known logical vulnerabilities specific to the super agent.

#### Enabling Learning and Error Prevention

- The guard agent **does not merely correct mistakes but provides deep, context-sensitive supervision**.
- By giving **stepwise, transparent diagnostics**, it ensures future reasoning by the super agent becomes more robust and less error-prone.
- The guard agent **fulfills a dual role**:
    1. **Quality control** of final answers.
    2. **Continuous improvement facilitator** for the super agent's performance, adapting feedback style based on a nuanced understanding of its design and historical output patterns.

### 4. Summary Table of Rounds

| Round | Actor        | Input Summary                         | Output Summary                                             |
| 1     | Super       | User question                         | Task structuring, tool preparation                        |
| 2     | Super       | Wikipedia query                       | Wikipedia article candidate found                         |
| 3     | Super       | Fetch Wikipedia content               | Article content extracted                                 |
| 4     | Guard       | Super agent's reasoning and candidate | Detailed diagnosis, error pinpointing, actionable feedback|
| 5     | Super       | Guard agent's advice                  | Formatted final answer submitted                          |
| 6     | System      | Super agent's final answer            | Automated marking                                         |

### 5. Conclusion

It is **precisely due to the guard agent's intimate awareness of the super agent's reasoning patterns, limitations, and strengths** that it can deliver **surgical feedback**, offering both corrective and developmental guidance. The **synergy between the two agents** ensures both high-quality task completion and a virtuous cycle of reasoning improvement, with the guard agent as the indispensable enabler of reliability and learning.
\end{lstlisting}